\def\paperlanguage{}
\pdfoutput=1 

\documentclass[letterpaper, 10 pt, conference]{ieeeconf}  

\usepackage{bm}
\usepackage{cite}
\usepackage{flushend}
\include{preamble}

\IEEEoverridecommandlockouts                              

\title{\LARGE \textbf
  {
    \switchlanguage%
    {%
      Exceeding the Maximum Speed Limit of the Joint Angle for the Redundant Tendon-driven Structures of Musculoskeletal Humanoids
    }%
    {%
      冗長な筋を有する筋骨格ヒューマノイドの\\最大関節角速度を超越する動作戦略
    }%
  }
}

\author{Kento Kawaharazuka$^{1}$, Yuya Koga$^{1}$, Kei Tsuzuki$^{1}$, Moritaka Onitsuka$^{1}$, Yuki Asano$^{1}$, Kei Okada$^{1}$,\\ Koji Kawasaki$^{2}$, and Masayuki Inaba$^{1}$
  \thanks{$^{1}$ The authors are with the Department of Mechano-Informatics, Graduate School of Information Science and Technology, The University of Tokyo, 7-3-1 Hongo, Bunkyo-ku, Tokyo, 113-8656, Japan.
    {\texttt\small [kawaharazuka, koga, tsuzuki, onitsuka, asano, k-okada, inaba]@jsk.t.u-tokyo.ac.jp}
  }
  \thanks{$^{2}$ The author is associated with TOYOTA MOTOR CORPORATION.
    {\texttt\small koji\_kawasaki@mail.toyota.co.jp}
  }
}
\begin{document}

\maketitle
\thispagestyle{empty}
\pagestyle{empty}

\begin{abstract}
  \switchlanguage%
  {%
    The musculoskeletal humanoid has various biomimetic benefits, and the redundant muscle arrangement is one of its most important characteristics.
    This redundancy can achieve fail-safe redundant actuation and variable stiffness control.
    However, there is a problem that the maximum joint angle velocity is limited by the slowest muscle among the redundant muscles.
    In this study, we propose two methods that can exceed the limited maximum joint angle velocity, and verify the effectiveness with actual robot experiments.
  }%
  {%
    筋骨格ヒューマノイドには生物規範型の様々な利点が存在し, その中でも冗長な筋配置は重要な特徴の一つである.
    この冗長な筋配置により冗長駆動・可変剛性が実現する一方, 多数の筋の中で最も速度の遅い筋に最大速度が規定されてしまうという問題がある.
    そこで本研究では, この規定された最大関節角速度よりもより速い速度を出すための手法を二種類考案し, 実機によって検証を行う.
  }%
\end{abstract}

\section{INTRODUCTION}\label{sec:introduction}
\switchlanguage%
{%
  The musculoskeletal humanoid \cite{nakanishi2013design, jantsch2013anthrob, kawaharazuka2019musashi} has many biomimetic benefits such as the radioulnar structure of the forearm \cite{kawaharazuka2017forearm}, the flexible spine \cite{osada2011planar}, and the scapula structure with a wide range of motion \cite{sodeyama2007scapula}.
  One of the most important characteristics among these benefits is redundant muscle arrangement.
  This enables fail-safe redundant actuation that can continuously move even if a few muscles are ruptured \cite{kawaharazuka2019longtime, kawaharazuka2020autoencoder}, and variable stiffness control using the redundancy and nonlinear elastic elements \cite{kawaharazuka2019longtime, nakanishi2011kenzoh}.
  On the other hand, there is a problem that high internal muscle tension or slack of antagonistic muscles can occur due to the model error.
  To solve the problem, antagonist inhibition control \cite{kawaharazuka2017antagonist}, dynamic modification of antagonistic relationships \cite{koga2019modification}, and muscle relaxation control \cite{kawaharazuka2019relax} have been developed so far.

  In this study, we handle a newly found problem of the redundant muscle arrangement.
  The maximum joint angle velocity is limited by the slowest muscle among the redundant muscles.
  We propose methods to exceed the limited maximum joint angle velocity and solve the problem.
  In other words, this becomes not a problem but a benefit, in which the robot can move faster than the limited joint angle velocity that we have thought was the maximum so far.

  To increase the joint angle velocity, optimization methods by software \cite{terasawa2016swing} have been developed so far.
  However, these optimizations cannot make use of the hardware characteristics of musculoskeletal humanoids.
  Also, regarding the axis-driven robots with variable stiffness mechanism, velocity maximization methods using the hardware have been developed \cite{haddadin2011velocity, chen2013velocity}.
  In this study, we propose simple methods to exceed the limited maximum joint angle velocity by making use of the redundant tendon-driven characteristics.

  This paper is organized as follows.
  In \secref{sec:musculoskeletal-humanoids}, we will explain the basic musculoskeletal structure and its problem.
  In \secref{sec:proposed}, we will propose two simple methods to exceed the limited joint angle velocity.
  In \secref{sec:experiments}, we will conduct experiments of two proposed methods using the musculoskeletal humanoid Musashi \cite{kawaharazuka2019musashi}.
  In \secref{sec:discussion}, we will compare the experimental results and discuss the advantages and disadvantages of these two methods.
}%
{%
  筋骨格ヒューマノイド\cite{nakanishi2013design, wittmeier2013toward, jantsch2013anthrob, asano2016kengoro, kawaharazuka2019musashi} (\figref{figure:musculoskeletal-humanoid})は前腕の橈骨尺骨構造\cite{kawaharazuka2017forearm}や柔軟な背骨\cite{osada2011planar}, 可動域の広い肩甲骨関節\cite{sodeyama2007scapula}等, 多くの生物規範型構造の利点を有する.
  その中でも最も重要な特徴の一つが冗長な筋配置である.
  これにより, 筋が一本切れても動き続ける冗長駆動\cite{kawaharazuka2019longtime}や, 非線形弾性要素と合わせた可変剛性制御\cite{kawaharazuka2019longtime, nakanishi2011kenzoh}が可能となる.
  この冗長な筋の拮抗配置はこれら多くの利点が存在する一方, モデル化の誤差によって, 拮抗筋間で大きな内力や緩みが発生するという問題があった.
  この問題に対して, これまで拮抗筋抑制制御\cite{kawaharazuka2017antagonist}や拮抗修正制御\cite{koga2019modification}, 筋弛緩制御\cite{kawaharazuka2019relax}等による内力・緩みの緩和が行われてきている.

  そして本研究は, 新たに発見したもう一つの問題点を扱うものである.
  これは, 冗長な多数の筋の関節に対する速度が様々であるため, 最も速度の遅い筋に関節角速度が規定されてしまうという現象である.
  本研究はこの問題を解決し, 規定された最大関節角速度よりも速い速度を出そうとする手法の提案である.
  これは言い換えれば, 問題点ではなく, これまで最大と思われていた規定された関節角速度よりも速い速度を出すことができるという, ある意味利点でもある.

  これまで関節角速度を速くするためには, \cite{terasawa2016swing}のようなソフトウェアにおける最適化が行われてきた.
  しかし, これらは拮抗腱駆動のハードウェアの特性を活かすことはできない.
  また, 可変剛性機能を持つ軸駆動型ロボットにおいては, その弾性変化を用いた速度最大化が成されている\cite{haddadin2011velocity, chen2013velocity}.
  本研究では, 冗長な筋を有する筋骨格構造の特性を用いて, 最大関節角速度を超えるシンプルな手法を提案する.

  本研究の構成は以下である.
  \secref{sec:musculoskeletal-humanoids}では筋骨格構造の基本的な特徴・特性について述べる.
  \secref{sec:proposed}では最大関節角速度を超えるための２つの提案手法について述べる.
  \secref{sec:experiments}では具体的に筋骨格ヒューマノイドMusashi\cite{kawaharazuka2019musashi}の左腕を用いて, ２つの提案手法をシミュレーション・実機に置いて実行し, その性能を確かめる.
  \secref{sec:discussion}ではシミュレーション・実験結果について考察し, 2つの提案手法を比較する.
}%


\section{Musculoskeletal Humanoids} \label{sec:musculoskeletal-humanoids}
\switchlanguage%
{%
  In this study, we generalize our explanation so that the complex musculoskeletal humanoids \cite{nakanishi2013design, jantsch2013anthrob, kawaharazuka2019musashi}, in which the moment arms of muscles to joints are not constant, can be handled.
  The simple tendon-driven robots such as \cite{hirose1991tendon, kobayashi1998tendon} can also be handled.
  Although we assume that the muscle actuator is an electric motor, we can also apply the principle of this study to pneumatically actuated robots.
}%
{%
  本研究は主に\cite{nakanishi2013design, wittmeier2013toward, jantsch2013anthrob, asano2016kengoro, kawaharazuka2019musashi}等の, 関節に対して筋のモーメントアームが一定では無い複雑な筋骨格ヒューマノイドを扱えるように一般化して説明するが, \cite{hirose1991tendon, kobayashi1998tendon}のようなモーメントアームが一定の腱駆動ロボットに対しても同様に適用可能である.
  また, 説明の都合上筋アクチュエータは空気圧ではなく, 電気アクチュエータを前提とするが, 空気圧型のロボットにも本研究で提案する原理は適用可能であると考える.
}%

\subsection{Basic Structure of the Musculoskeletal Humanoid} \label{subsec:basic-structure}
\begin{figure}[t]
  \centering
  \includegraphics[width=0.70\columnwidth]{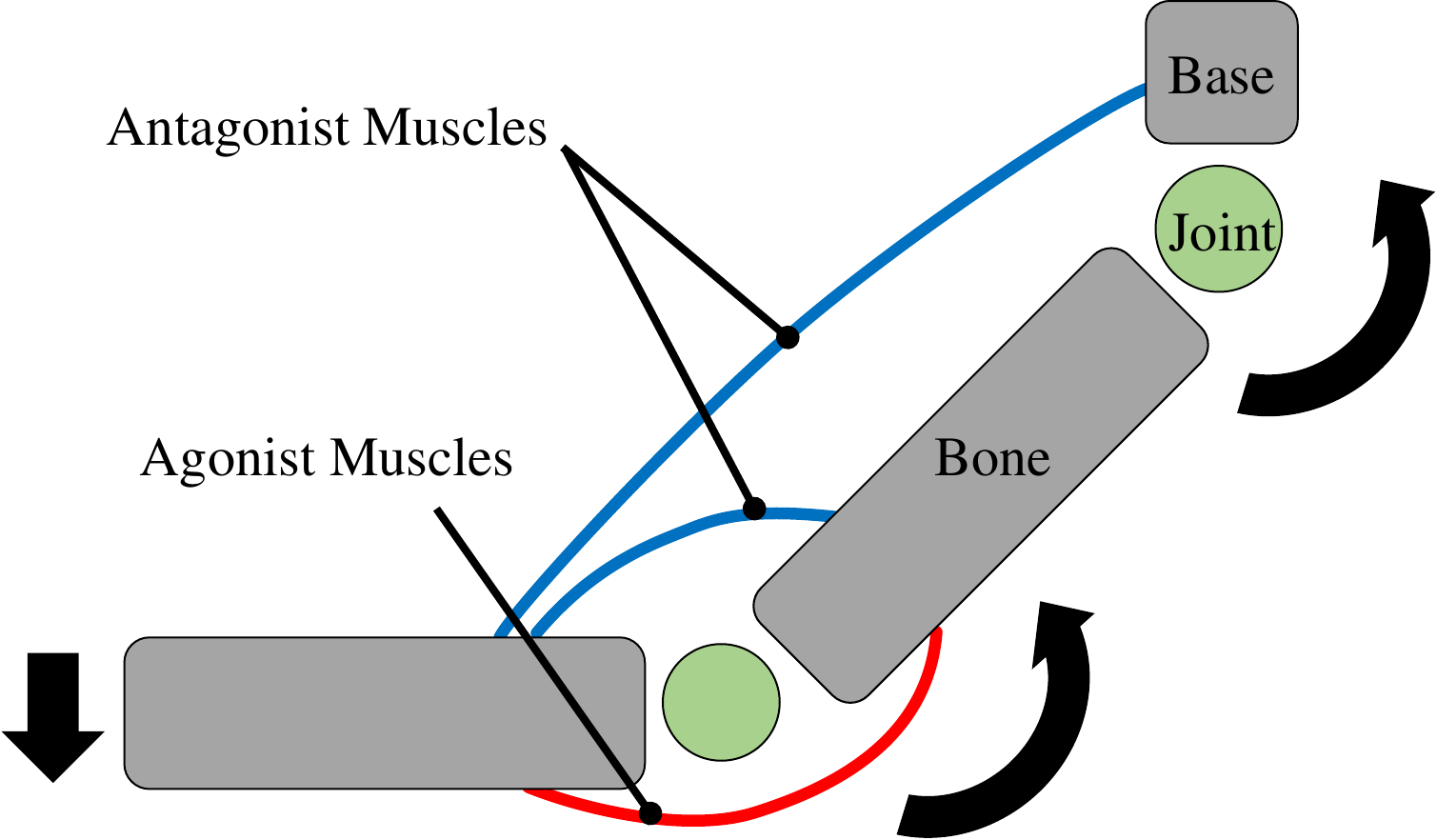}
  \caption{The basic musculoskeletal structure.}
  \label{figure:musculoskeletal-structure}
  \vspace{-3.0ex}
\end{figure}

\switchlanguage%
{%
  We show the basic musculoskeletal structure in \figref{figure:musculoskeletal-structure}.
  Muscles are antagonistically arranged around joints.
  The robot usually has not only monoarticular but also polyarticular muscles for the benefits of balancing and joint coordination \cite{sharbafi2016biarticular}.
  We call the muscles contributing to the direction of the intended movement ``agonist muscles,'' and the muscles restraining the movement ``antagonist muscles.''
  The relationship between joint angle and muscle length is represented as below,
  \begin{align}
    \bm{l} &= \bm{h}(\bm{\theta}) \label{eq:h}\\
    \Delta\bm{l} &= G(\bm{\theta})\Delta\bm{\theta} \label{eq:G}
  \end{align}
  where $\bm{l}$ is muscle length, $\bm{\theta}$ is joint angle, $\Delta\{\bm{l}, \bm{\theta}\}$ is a small displacement of $\{\bm{l}, \bm{\theta}\}$, $\bm{h}$ is a mapping from $\bm{\theta}$ to $\bm{l}$, and $G$ is muscle Jacobian which is a differential matrix of $\bm{h}$.
  Here, $\bm{l}$ is a $m$-dimensional vector and $\bm{\theta}$ is a $n$-dimensional vector ($m$ and $n$ are the numbers of muscles and joints, respectively).
}%
{%
  一般的な筋骨格構造を\figref{figure:musculoskeletal-structure}に示す.
  関節の周りに筋が冗長に配置されている.
  そのバランスや関節協調の利点\cite{sharbafi2016biarticular}ゆえに, 一つの関節を動かす単関節筋だけでなく, 多関節筋も有している場合がほとんどである.
  これらの筋を動作させる筋アクチュエータは全て同じとは限らず, それぞれギア比やワット数が異なることがある.
  動作させる方向に寄与する筋は主動筋, 阻害する方向に寄与する筋は拮抗筋と呼ばれる.
  関節角度と筋長の関係は一般的に以下のように表される.
  \begin{align}
    \bm{l} &= \bm{h}(\bm{\theta}) \label{eq:h}\\
    \Delta\bm{l} &= G(\bm{\theta})\Delta\bm{\theta} \label{eq:G}
  \end{align}
  ここで, $\bm{l}$は筋長, $\bm{\theta}$は関節角度, $\Delta\{\bm{l}, \bm{\theta}\}$は$\{\bm{l}, \bm{\theta}\}$の微少変位, $\bm{h}$は$\bm{\theta}$のときの$\bm{l}$を表す写像, $G$は$\bm{h}$の微分である筋長ヤコビアンを表す.
  これらの関係は, 人間が筋の起始点・中継点・終止点を結んで作成した幾何モデルと実機の間では大きな誤差があり, \cite{kawaharazuka2018online, kawaharazuka2019longtime}等の手法で実機のセンサ情報から学習させることが望ましい.
}%

\subsection{Problem of the Redundant Tendon-driven Structure Addressed in This Study} \label{subsec:basic-problem}
\begin{figure}[t]
  \centering
  \includegraphics[width=0.7\columnwidth]{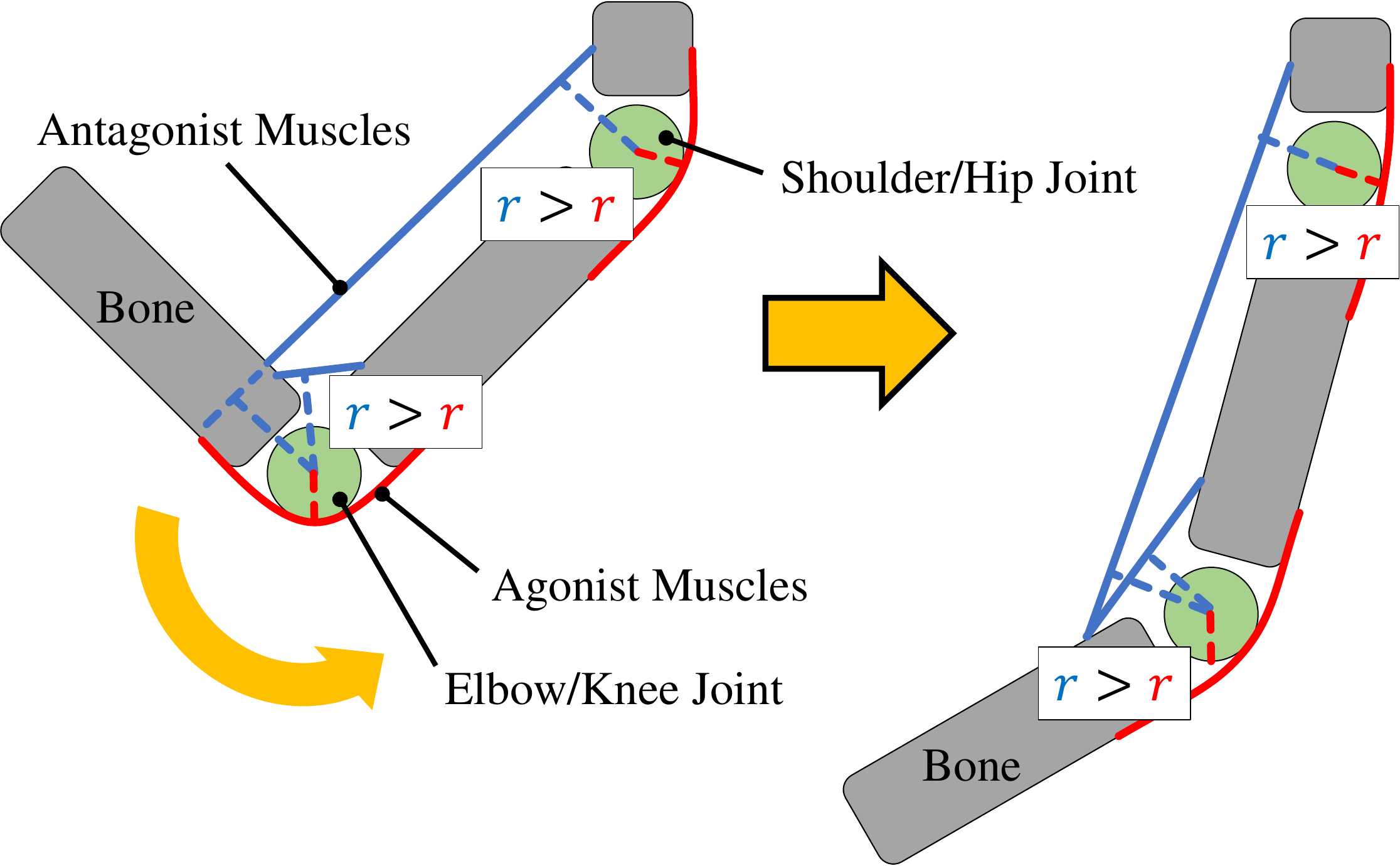}
  \caption{Characteristics of movements with high joint velocity. These movements use the effect of gravity and body inertia.}
  \label{figure:motion-primitives}
  \vspace{-3.0ex}
\end{figure}

\switchlanguage%
{%
  As a movement with fast joint angle velocity, we consider the movement of swinging down the arm, such as striking the desk, hitting with a hammer, and swinging a golf club.
  We determine the joint angle $\bm{\theta}^{start}$ when the arm is swung up, and the joint angle $\bm{\theta}^{end}$ when swung down.
  If we represent the current joint angle as $\bm{\theta}$ and the current joint angle velocity as $\dot{\bm{\theta}}$ during the motion of swinging down the arm, the moment arms of muscles to joints $\bm{r}$ are represented as below,
  \begin{align}
    \bm{r}(\bm{\theta}, \dot{\bm{\theta}}) &= G(\bm{\theta})\dot{\bm{\theta}}/||\dot{\bm{\theta}}||_{2} \label{eq:moment-arm}
  \end{align}
  where $||\cdot||_{2}$ represents L2 norm.
  If $\bm{r}$ of the muscle is positive, it is an antagonist muscle, and if $\bm{r}$ of the muscle is negative, it is an agonist muscle.
  The absolute value of $\bm{r}$ is the moment arm.
  If the muscle is a polyarticular muscle involving multiple joints or a muscle whose position is distant from joints, the moment arm becomes large.
  The larger the moment arm is, the faster the muscle length velocity must be, even with the same joint angle velocity.
  A muscle with a large moment arm can easily achieve the maximum muscle length velocity of the actuator limit $\dot{\bm{l}}^{limit}$.
  Thus, the larger the index $\bm{q}$ as below is, the easier $\dot{\bm{l}}^{limit}$ is achieved.
  \begin{align}
    \bm{q} = \bm{r}((\bm{\theta}^{start}+\bm{\theta}^{end})/2, \bm{\theta}^{end}-\bm{\theta}^{start})\oslash\dot{\bm{l}}^{limit}
  \end{align}
  where $\oslash$ is the element-wise division of the vector.
  $\bm{q}$ uses the muscle Jacobian at the center of $\bm{\theta}^{start}$ and $\bm{\theta}^{end}$.

  Also, muscles that can easily achieve $\dot{\bm{l}}^{limit}$ have another characteristic besides the large moment arms and the small $\dot{\bm{l}}^{limit}$.
  When considering the movements with fast joint angle velocity, many movements such as hammer hitting, soccer, and golf use gravity.
  Thus, we move the body using not only the actuator output but also the effects of gravity and body inertia.
  We show the characteristics of such movements in \figref{figure:motion-primitives} (this can be interpreted as the movement of the shoulder-elbow or hip-knee).
  In this situation, when comparing the moment arms, those of antagonist muscles are always larger than those of agonist muscles.
  This is because while antagonist muscles bridge the bones, agonist muscles pass along joints.
  Therefore, in the case of the same $\dot{\bm{l}}^{limit}$, antagonist muscles are slower than agonist muscles, and the joint angle velocity is restrained even if agonist muscles move fast.
  If these antagonist muscles do not hinder the movement, the robot can move at the same speed as agonist muscles but also at a faster speed than the agonist muscles because of the effects of gravity and body inertia.

  In addition, regarding agonist muscles, even if one of them is slower than the others, it does not restrain the movement of joints.
  This fact is always true, even though the situation of \figref{figure:motion-primitives} does not occur when moving against gravity.
  Thus, the main cause limiting the joint angle velocity is the movement of antagonist muscles.

  In this study, we focus on the management of antagonist muscles with large $\bm{q}$ so that they do not achieve $\dot{\bm{l}}^{limit}$.
}%
{%
  速い関節角速度を出す動作として, 手を振り下ろす動作を考えてみる.
  例として, 机を叩いたり, ハンマーを振り下ろすような動作を考えてもらえると良い.
  このとき, 手を振り上げたとき関節角度$\bm{\theta}^{start}$と振り下ろす目標の関節角度$\bm{\theta}^{end}$を決める.
  手を振り下ろしている最中の現在の関節角度を$\bm{\theta}$, このときの関節角速度(現在関節角度から目標関節角度方向へのベクトル)を$\dot{\bm{\theta}}$とすると, 現在の動作方向に関する筋のモーメントアーム$\bm{r}$は以下のようになる.
  \begin{align}
    \bm{r}(\bm{\theta}) &= G(\bm{\theta})\dot{\bm{\theta}}/||\dot{\bm{\theta}}||_{2} \label{eq:moment-arm}
  \end{align}
  ここで, $||\cdot||_{2}$はL2ノルムを表す.
  $\bm{r}$が正なら拮抗筋, $\bm{r}$が負なら主動筋を表す.
  この$\bm{r}$の絶対値はモーメントアームであり, 複数の関節を跨ぐ多関節筋や関節から筋の位置までが遠い筋では大きな値となる.
  そして, このモーメントアームが大きいほど同一の関節角速度でも筋長速度が大きくなり, アクチュエータの限界である最大筋長速度に達しやすいことになる.
  実際に最大筋長速度に達するかどうかは, 筋アクチュエータごとの最大筋長速度も考慮する必要がある.
  よって, 以下の指標$\bm{q}$が大きな筋ほど最大筋長速度に達しやすい.
  \begin{align}
    \bm{q}(\bm{\theta}) = \bm{r}(\bm{\theta})\oslash\dot{\bm{l}}^{limit}
  \end{align}
  ここで, $\oslash$はベクトルの要素ごとの除算, $\dot{\bm{l}}^{limit}$は最大筋長速度を表す.

  また, これらの最大筋長速度に達しやすい筋には, モーメントアームが大きい・最大筋長速度が小さいということ以外にも特徴がある.
  速い関節角速度を出すような動きを考えてみると, 例えば手を振り下ろしたり, サッカーボールを蹴ったり等, 重力の影響を使ったものが多い.
  つまり, アクチュエータの速度だけでなく, 重力や身体の慣性を活かして動作させている.
  ゆえに, それらの動作は上から下へ動く, つまり, \figref{figure:motion-primitives}のような動作がほとんどである(これは肩と肘の動きとしても, 股と膝の動きとしても取れる).
  このとき, 筋のモーメントアームを見てみると, どの動作においても, 動作を阻害する方向の筋, つまり拮抗筋のモーメントアームが主動筋よりも大きくなっていることがわかる.
  これは, 拮抗筋では筋が骨格間を張っているのに対して, 主動筋では筋が関節を張っているためである.
  ゆえに, 最大筋長速度が同じ場合, これらの動作では主に拮抗筋の方が遅いため, 主動筋がどんなに頑張っても, その動きが阻害されて, 速度が制限されてしまうのである.
  もしここで拮抗筋が働かなければ, より速い速度で主動筋が関節を動かすだけでなく, 重力や慣性の力によって, 主動筋の速度よりも速く関節が動く可能性がある.
  また, 主動筋は一部が遅くても関節の動作を阻害することはなく, 拮抗筋が関節角速度を制限する主な理由であることがわかる.
  つまり, \figref{figure:motion-primitives}のような動作でなくとも, 基本的には拮抗筋が関節角速度を阻害する主な理由となる.

  よって, 値$\bm{q}$が大きな拮抗筋を何らかの方法で最大筋長速度まで達しないように管理することが, 本研究の主眼である.
}%

\section{A Method Exceeding the Limited Speed of the Joint Angle} \label{sec:proposed}
\switchlanguage%
{%
  To solve the problem of \secref{subsec:basic-problem}, we will propose two simple methods.
  The first one is a method inhibiting antagonist muscles with large $\bm{q}$, thus making their current 0, and using backdrivability of muscles.
  The second one is a method elongating antagonist muscles with large $\bm{q}$ in advance.
}%
{%
  \secref{subsec:basic-problem}の問題を解決するために, 本研究では二つの手法について考案・実験・考察をしていく.
  一つ目はモーメントアームの大きな拮抗筋を抑制, つまり流れる電流を0とすることで, バックドライバビリティを使って筋を出すという手法である.
  二つ目はモーメントアームの大きな拮抗筋を予め緩めておくという手法である.
}%

\subsection{Method Inhibiting Antagonist Muscles} \label{subsec:proposed-backdrivability}
\switchlanguage%
{%
  This is a method not to manage the antagonist muscles but to inhibit them.
  This control is very simple, and the muscle motor current with large $\bm{q}$ is inhibited to 0 as below, soon after starting the movement,
  \begin{align}
    \bm{o}[i] = 0\;\;\;if\;\bm{q}[i] > C \label{eq:backdrivability}
  \end{align}
  where $i$ is the muscle index, $\bm{o}$ is the current of the motor, and $C$ is a constant value.
  If $C=\bm{0}$, the currents of all the antagonist muscles become $\bm{0}$, and if $C>0$, the currents of antagonist muscles with large moment arms and small $\dot{\bm{l}}^{limit}$ become $\bm{0}$.
  If the muscle actuators have backdrivability, the muscles spontaneously elongate when pulled, and we do not have to consider their maximum muscle length velocities.
  Because $\bm{q}$ gradually changes during the movement, this control runs at a high frequency.
  This method assumes that the muscle actuators have backdrivability, and we verify it later using the actual robot.
}%
{%
  最大筋長速度まで達しないように管理するのではなく, 逆にそれを管理せずに出力を0にしてしまうのが本節の手法である.
  この手法は非常に単純であり, 動作開始すぐに, 以下のように$\bm{q}$が大きな筋の出力を0にしてしまえば良い.
  \begin{align}
    \bm{o}[i] = 0\;\;\;if\;\bm{q}[i] > C \label{eq:backdrivability}
  \end{align}
  ここで, $i$は着目する筋の番号, $\bm{o}$は電流値, $C$は定数である.
  $C$を0とすると, 拮抗筋のみの電流値が0となり, $C>0$とすると, 拮抗筋の中でもモーメントアームが大きな筋のみの電流値が0となる.
  もしその筋にバックドライバビリティがあれば, 筋は引っ張られることで自然と出ていき, それらの筋には最大筋長速度が存在しなくなる.
  $\bm{q}$の大きさは動作中に徐々に変化していくため, この制御は常に低レイヤーで走り続ける.
  この手法はバックドライバビリティがあることが前提となるが, これは実機験において検証する.
}%

\subsection{Method Elongating Antagonist Muscles} \label{subsec:proposed-elongation}
\switchlanguage%
{%
  This is a method managing the maximum muscle velocity by elongating the antagonist muscles with large moment arms in advance.
  When choosing which muscle we should elongate, we must consider two points stated below.

  First, when swinging up the arm or leg, the posture of $\bm{\theta}^{start}$ must be achievable.
  Because antagonist muscles when swinging down are agonist muscles when swinging up, if we simply elongate them when swung up, the robot cannot achieve the posture of $\bm{\theta}^{start}$.
  Thus, when considering a mask $\bm{m}$ whose values of muscles to elongate are $\bm{0}$ and those not to elongate are $\bm{1}$, the quadratic programming below must have a solution.
  \begin{align}
    \underset{\bm{f}}{\textrm{minimize}}&\;\;\;\;\;\;\;\;\;\;\;\;\;\;\;\;\;(\bm{m}\otimes\bm{f})^{T}W_{1}(\bm{m}\otimes\bm{f})\label{eq:balance}\\
    \textrm{subject to}&\;\;\;\;\;\;\;\;\;\;\;\; \bm{\tau}^{nec} = -G^{T}(\bm{\theta}^{start})(\bm{m}\otimes\bm{f})\nonumber\\
    &\;\;\;\;\;\;\;\;\;\;\;\; \bm{m}\otimes\bm{f}^{min} \leq \bm{m}\otimes\bm{f} \leq \bm{m}\otimes\bm{f}^{max}\nonumber
  \end{align}
  where $\otimes$ is element-wise multiplication, $\bm{f}$ is the calculated muscle tension, $W_{1}$ is a weight matrix (identity matrix in this study), $\bm{\tau}^{nec}$ is a joint torque which is necessary to achieve $\bm{\theta}^{start}$, and $\bm{f}^{\{min, max\}}$ is a minimum or maximum muscle tension.
  If $\bm{f}$ satisfying this condition can be calculated, $\bm{\theta}^{start}$ is achievable, and the robot can elongate the muscles whose $\bm{m}$ is 0 while keeping $\bm{\theta}^{start}$.
  This is a principle enabled by the muscle redundancy.
  Although elongating the muscles increases the muscle tension of the others, it is not a problem for a short time.

  Second, by elongating the chosen antagonist muscles, the joint angle velocity must become faster.
  By determining $\bm{m}$ and conducting a simulation as below, we can calculate the time cost to achieve $\bm{\theta}^{start}$ to $\bm{\theta}^{end}$,
  \begin{align}
    \underset{\Delta\bm{\theta}}{\textrm{minimize}}&\;\;\;(\bm{\theta}^{end}-\bm{\theta}-\Delta\bm{\theta})^{T}W_{2}(\bm{\theta}^{end}-\bm{\theta}-\Delta\bm{\theta})\label{eq:simulation}\\
    \textrm{subject to}&\;\;\; -\bm{m}\otimes\dot{\bm{l}}^{limit}\Delta{t} \leq \bm{m}\otimes(G(\bm{\theta})\Delta\bm{\theta}) \leq \bm{m}\otimes\dot{\bm{l}}^{limit}\Delta{t}\nonumber
  \end{align}
  where $\Delta\bm{\theta}$ is the simulated displacement of the joint angle from $\bm{\theta}$, $W_{2}$ is a weight matrix (identity matrix in this study), $t$ is the current time step, and $\Delta{t}$ is the time interval of simulation.
  We can calculate $\Delta\bm{\theta}$ representing how much $\bm{\theta}$ can get closer to $\bm{\theta}^{end}$ in $\Delta{t}$ seconds.
  We update this simulation like $\bm{\theta}\gets\bm{\theta}+\Delta\bm{\theta}$ and $t{\gets}t+\Delta{t}$, by starting from $\bm{\theta}=\bm{\theta}^{start}$.
  We stop the simulation when $||\bm{\theta}-\bm{\theta}^{end}||_{2}<\epsilon$, and the last $t$ is the time cost of the movement $t^{cost}$.
  We need to calculate $\bm{m}$ that makes $t^{cost}$ smaller.
  Although this calculation is a rough estimate because this simulation does not consider the motor inertia, model error, hysteresis, etc., we can obtain the rough characteristics of the movement.
  Because the musculoskeletal humanoid is difficult to modelize due to its complex structure compared with the ordinary axis-driven humanoid, we use such a simple method.

  We search $\bm{m}$ which can achieve $\bm{\theta}^{start}$ by the calculated muscle tension and which makes $t^{cost}$ smaller.
  Although we can conduct a full search of all the candidates of $\bm{m}$, $\bm{t}^{cost}$ clearly decreases when not using muscles with large $\bm{q}$.
  Therefore, we make $\bm{m}$ of antagonist muscles equal $\bm{0}$ in decreasing order of $\bm{q}$, and stop the search when $\bm{f}$ achieving $\bm{\theta}^{start}$ no longer exists.

  Finally, we calculate how long the muscles, whose $\bm{m}$ are $\bm{0}$, should be elongated.
  We can obtain the transition of $\Delta\bm{\theta}$ using a simulation conducted with the calculated $\bm{m}$.
  From the transition of joint angle, we can calculate the transition of muscle length using $G(\bm{\theta})\Delta\bm{\theta}$.
  The maximum difference between the calculated muscle length transition and the fastest muscle length transition to elongate the muscles by $\dot{\bm{l}}^{limit}$ is $\Delta\bm{l}^{elongate}$, which is the minimum amount of muscle length that should be elongated.
  By elongating the chosen muscles by $\Delta\bm{l}^{elongate}$ at $\bm{\theta}=\bm{\theta}^{start}$ in advance, the antagonist muscles do not restrain the movement of agonist muscles, and the robot can move faster.
}%
{%
  モーメントアームの大きな拮抗筋を予め緩めることで, 確実に最大筋長速度まで達しないように管理するのが本節の手法である.
  どの筋を予め緩めるかを考える際に, 以下の2つの点を考慮する必要がある.

  まず, 手や足を振り上げるときに, その, 重力に逆らった姿勢が実現可能でなければならない.
  動作方向, つまり手や足を振り下げる動作の拮抗筋は振り上げる動作の主動筋であり, これらを単純に緩めてしまっては, その姿勢を実現することができない.
  つまり, 予め緩める筋を0, 緩めない筋を1としたマスクベクトル$\bm{m}$を考えたときに, 以下の二次計画法において解がある, つまり姿勢$\bm{\theta}^{start}$を実現できることを保証しなければならない.
  \begin{align}
    \underset{\bm{f}}{\textrm{minimize}}&\;\;\;\;\;\;\;\;\;\;\;\;\;\;\;\;\;(\bm{m}\otimes\bm{f})^{T}W_{1}(\bm{m}\otimes\bm{f})\label{eq:balance}\\
    \textrm{subject to}&\;\;\;\;\;\;\;\;\;\;\;\; \bm{\tau}^{nec} = -G^{T}(\bm{\theta}^{start})(\bm{m}\otimes\bm{f})\nonumber\\
    &\;\;\;\;\;\;\;\;\;\;\;\; \bm{m}\otimes\bm{f}^{min} \leq \bm{m}\otimes\bm{f} \leq \bm{m}\otimes\bm{f}^{max}\nonumber
  \end{align}
  ここで, $\otimes$はベクトルの要素ごとの乗算, $\bm{f}$は計算される筋張力, $W_{1}$は重み行列(本研究では単位行列とする), $\bm{\tau}^{nec}$は$\bm{\theta}^{start}$を実現するために必要な関節トルク, $\bm{f}^{\{min, max\}}$は筋張力の最小値・最大値を表す.
  これを満たす筋張力$\bm{f}$が計算可能であれば, $\bm{\theta}^{start}$は実現可能であり, $\bm{\theta}^{start}$のまま$\bm{m}$で指定した筋を緩ませることが可能である.
  これは筋の冗長性ゆえに可能となる原理である.
  筋を緩めることで他の筋に対する張力が増えてしまうが, 少しの時間であれば問題ないと考え, このような動作戦略を取った.

  次に, 当然その拮抗筋を緩めることで, 関節角速度が速くならなければならない.
  $\bm{m}$を決め, 以下のようなシミュレーションを行うことで, 目標関節角度まで移動する時間を概算することができる.
  \begin{align}
    \underset{\Delta\bm{\theta}}{\textrm{minimize}}&\;\;\;(\bm{\theta}^{end}-\bm{\theta}-\Delta\bm{\theta})^{T}W_{2}(\bm{\theta}^{end}-\bm{\theta}-\Delta\bm{\theta})\label{eq:simulation}\\
    \textrm{subject to}&\;\;\; -\bm{m}\otimes\dot{\bm{l}}^{limit}\Delta{t} \leq \bm{m}\otimes(G(\bm{\theta})\Delta\bm{\theta}) \leq \bm{m}\otimes\dot{\bm{l}}^{limit}\Delta{t}\nonumber
  \end{align}
  ここで, $\Delta\bm{\theta}$は現在関節角度$\bm{\theta}$からの予想される変位, $W_{2}$は重み行列(本研究では単位行列とする), $t$は現在時刻, $\Delta{t}$はシミュレーションの間隔を表す.
  これにより, $\bm{\theta}$から$\Delta{t}$秒間で最大どの程度$\bm{\theta}^{end}$に近づけるかを表す$\Delta\bm{\theta}$を計算することができる.
  このシミュレーションを$\bm{\theta}=\bm{\theta}^{start}$からはじめ, $\bm{\theta}\gets\bm{\theta}+\Delta\bm{\theta}$, $t{\gets}t+\Delta{t}$というように更新していく.
  $||\bm{\theta}-\bm{\theta}^{end}||_{2}<\epsilon$となったところでシミュレーションを終了し, そのときの$t$がかかった時間$t^{cost}$となる.
  この$t^{cost}$が小さくなるような$\bm{m}$を求める必要がある.
  実際にはロボット・モータの慣性や筋経路のモデル誤差, 摩擦等が多く存在するため, この計算はあくまで簡易的な概算であるが, それでも大まかな特性を知ることが可能である.
  軸駆動型ロボットと違い筋骨格ヒューマノイドはその複雑な構造ゆえに詳細なモデル化が難しいため, このようなシンプルな方法を取った.

  上記の$\bm{\theta}^{start}$を実現できる筋張力が存在し, $t^{cost}$が小さいような$\bm{m}$を探索していく.
  全探索を行っても良いが, $\bm{q}$が大きな筋を抜くことで$\bm{t}^{cost}$が下がることは明白である.
  そのため, $\bm{q}$が大きな筋から順に$\bm{m}$の値を0にしていき, $\bm{\theta}^{start}$を実現できる$\bm{f}$が存在しなくなったところで探索をやめる, という手法を取る.

  最後に, $\bm{m}$の値が0となった筋をどの程度伸ばす必要があるかを計算する.
  最終的に求まった$\bm{m}$によって先のシミュレーションを行って$\Delta\bm{\theta}$の遷移を得る.
  この遷移から$G(\bm{\theta})\Delta\bm{\theta}$によって筋長の遷移が計算でき, 常に最大筋長速度で筋を伸ばした時の筋長遷移との差分の最大値が, 伸ばすべき最小の筋長変化$\Delta\bm{l}^{elongate}$となる.
  $\bm{\theta}^{start}$の時点で$\Delta\bm{l}^{elongate}$だけ筋長を伸ばすことで, 拮抗筋が主動筋を阻害せず, より速い動作が可能となる.
}%

\begin{figure}[t]
  \centering
  \includegraphics[width=0.9\columnwidth]{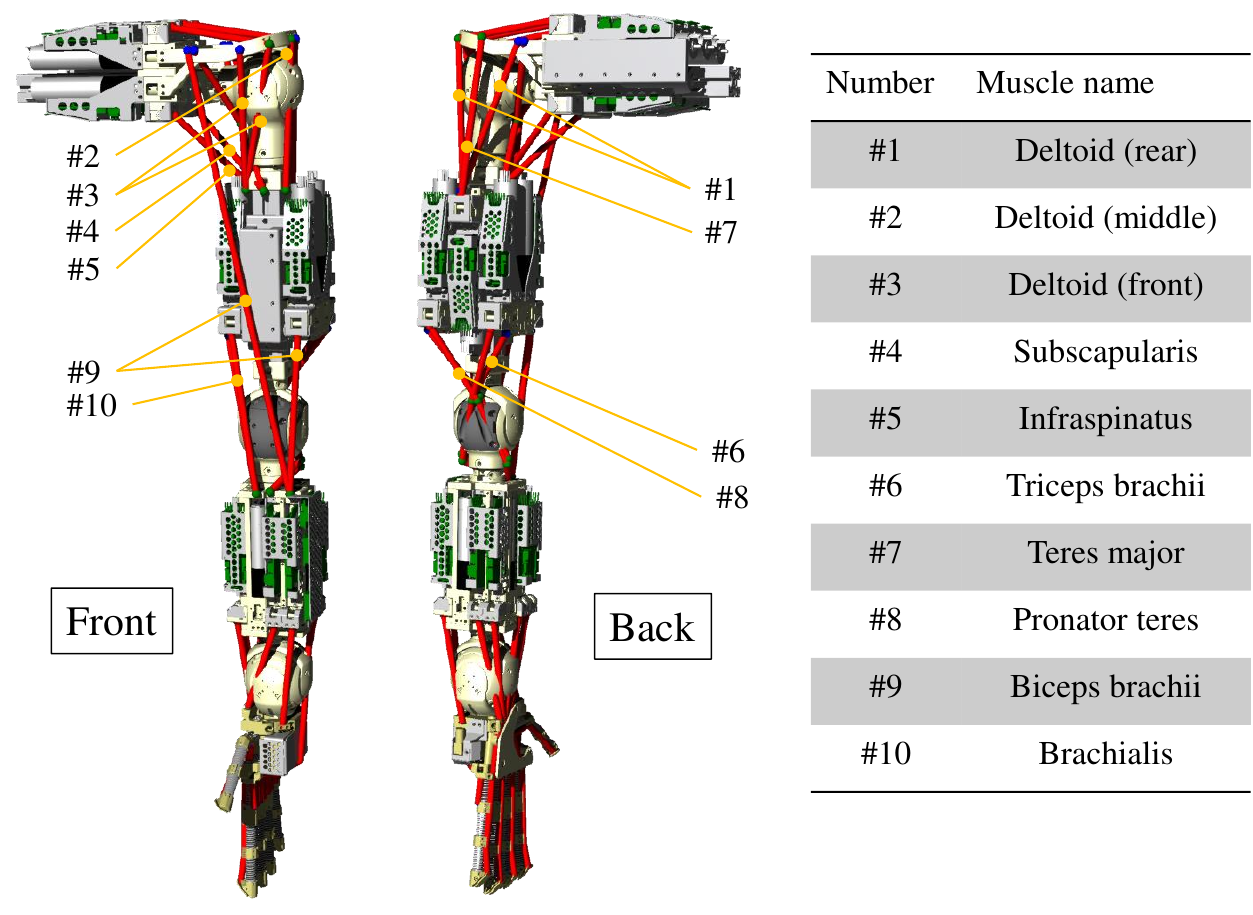}
  \caption{Muscle arrangement of the left arm of the musculoskeletal humanoid Musashi \cite{kawaharazuka2019musashi}.}
  \label{figure:muscle-arrangement}
  \vspace{-1.0ex}
\end{figure}

\begin{figure}[t]
  \centering
  \includegraphics[width=0.8\columnwidth]{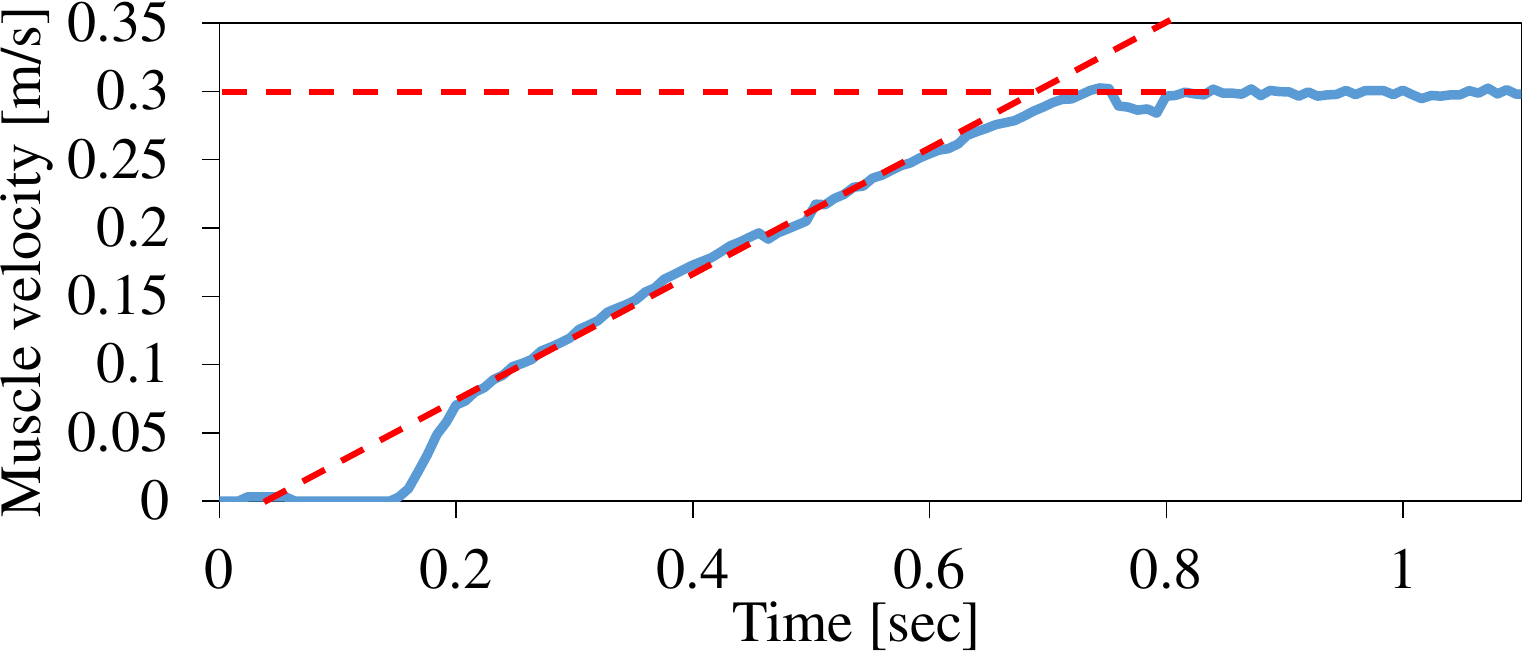}
  \caption{Characteristics of muscle length velocity transition.}
  \label{figure:muscle-characteristic}
  \vspace{-3.0ex}
\end{figure}

\begin{figure}[t]
  \centering
  \includegraphics[width=0.8\columnwidth]{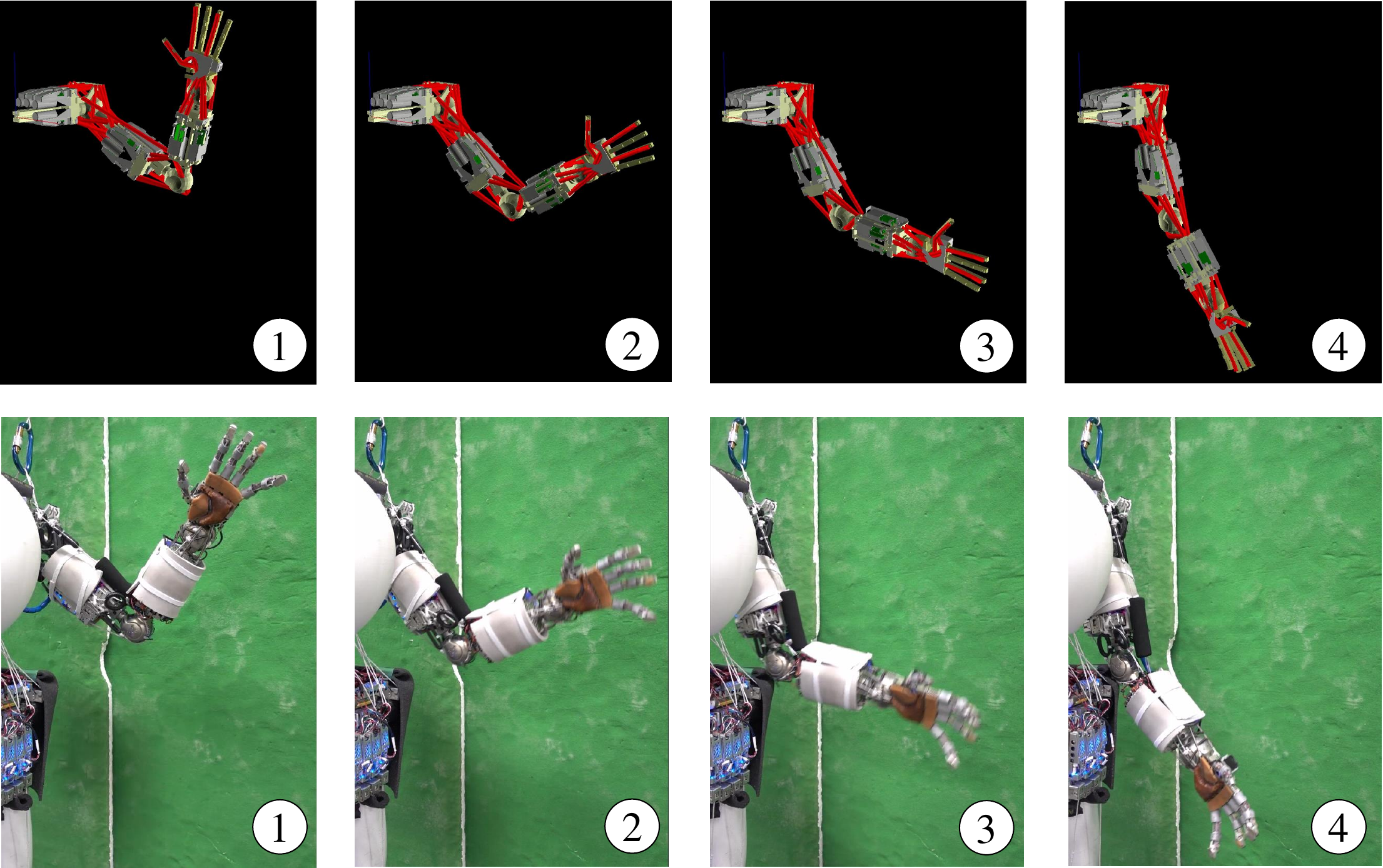}
  \caption{Experimental motion of simulation and actual robot.}
  \label{figure:experimental-motion}
  \vspace{-3.0ex}
\end{figure}

\section{Experiments} \label{sec:experiments}
\subsection{Experimental Setup}
\switchlanguage%
{%
  In this study, we use the left arm of the musculoskeletal humanoid Musashi \cite{kawaharazuka2019musashi} for experiments.
  We show its muscle arrangement in \figref{figure:muscle-arrangement}.
  We mainly use its five DOFs of the shoulder and elbow.
  We represent these joint angles as S-p, S-r, S-y, E-p, E-y (S is the shoulder, E is the elbow, and rpy is roll, pitch, and yaw).
  These joints involve ten muscles including one polyarticular muscle.
  The motors of all the muscle actuators \cite{asano2015sensordriver} are 90W Maxon BLDC Motor with 29:1 gear ratio, and $\dot{\bm{l}}^{limit}$ of them are the same.
  However, the current control in \cite{asano2015sensordriver} cannot achieve $\dot{\bm{l}}^{limit}$ quickly.
  This is because the motor driver uses single-shunt approach to reduce the substrate size and we cannot increase the gain of current control to avoid the vibration of output.
  Therefore, we replace $-\dot{l}^{limit}$ and $\dot{l}^{limit}$ in \equref{eq:simulation} by $\dot{l}^{min}$ and $\dot{l}^{max}$, and we update them as below when the current muscle length velocity $\dot{l}>0$,
  \begin{align}
    \dot{l}^{max} &= \textrm{min}(\dot{l}+\alpha\Delta{t}, \dot{l}^{limit})\\
    \dot{l}^{min} &= \textrm{min}(\dot{l}-\alpha\Delta{t}, 0)
  \end{align}
  where $\alpha$ is a constant value.
  In this study, from \figref{figure:muscle-characteristic}, we identified that $\alpha=0.46$ [m/s$^2$] and $\dot{l}^{limit}=0.30$ [m/s].
  This is a constraint that while $\dot{\bm{l}}$ can gradually increase in proportion to time, $\dot{\bm{l}}$ can decrease to $\bm{0}$ at once.
  When $\dot{l}<0$, the constraint is the same that $\dot{\bm{l}}$ can gradually decrease in proportion to time, and $\dot{\bm{l}}$ can increase to $\bm{0}$ at once.
  This expressed the behavior of actual muscle modules well.

  In this study, we handle the movement of swinging down the left arm of Musashi.
  We show the experimental movements in simulation and in the actual robot, in \figref{figure:experimental-motion}.
  Although we cannot measure the joint angle of the ordinary musculoskeletal humanoid due to the complex joint structures, we can measure the joint angle of Musashi using the equipped joint modules.
  Also, we can measure muscle length from the encoder attached to the muscle actuator and muscle tension from the tension measurement unit.
  In this study, we set $C=0$, $\Delta{t}=0.03$, $\bm{f}^{min}=10$ [N], and $\bm{f}^{max}=200$ [N].
}%
{%
  本研究では, 筋骨格ヒューマノイドMusashi \cite{kawaharazuka2019musashi}の左腕を用いる.
  その筋配置は\figref{figure:muscle-arrangement}のようになっている.
  本研究では主に肩と肘の5自由度を用い, それらに関係する筋は10本, うち二関節筋が1本含まれている.
  これらの関節角度をS-p, S-r, S-y, E-p, E-yと表す(Sは肩関節, Eは肘関節, rpyはroll, pitch, yawを表す).
  全筋アクチュエータ\cite{asano2015sensordriver}のモータは90W, ギア比29:1のMaxon BLDC Motorであり, 最大筋長速度の特性は一致している.
  ただし, \cite{asano2015sensordriver}における目標筋長と現在筋長の差分に対して電流を流すような制御では\figref{figure:muscle-characteristic}に示すように最大筋長速度に一瞬で到達するわけではない.
  これは, モータドライバを小さくするために1シャント方式を採用しており, 出力が振動するためゲインを上げられないことが原因であると考えられる.
  そのため, \secref{subsec:proposed-elongation}の\equref{eq:simulation}における$-\dot{l}^{limit}$と$\dot{l}^{limit}$を$\dot{l}^{min}$と$\dot{l}^{max}$に置き換え, 現在の筋長が$\dot{l}>0$のとき, 以下のようにそれらを更新して用いる.
  \begin{align}
    \dot{l}^{max} &= \textrm{min}(\dot{l}+\alpha\Delta{t}, \dot{l}^{limit})\\
    \dot{l}^{min} &= \textrm{min}(\dot{l}-\alpha\Delta{t}, 0)
  \end{align}
  ここで, $\alpha$は定数であり, 本研究では, \figref{figure:muscle-characteristic}から$\alpha=0.46$ [m/s$^2$], $\dot{l}^{limit}=0.30$ [m/s]と同定された.
  これは, 現在筋長速度を現在から上げる場合は時間に比例して徐々に速度が上がるのに対して, 下がる場合は一気に0まで下げることができる, という制約である.
  $\dot{l}<0$の場合も同様で, より速いマイナス方向の速度を出すには制限がかかるが, 速度を0まで戻すことに対して制限はない.

  本研究ではMusashiが左手を振り下ろすような動作について扱う.
  シミュレータ, 実機における動きは\figref{figure:experimental-motion}のようになっている.
  このときの関節角度, 筋長の速度等を記録していく.
  通常の筋骨格ヒューマノイドは肩甲骨や球関節等の複雑な関節により関節角度を測定することができないが, 本研究で扱うMusashiは関節モジュール内に角度センサを有するため, これを測定することが可能である.
  また, 筋長は筋アクチュエータについたエンコーダの値を用いている.
  本研究のパラメータとして, $C=0$, $\Delta{t}=0.03$とする.
}%

\subsection{Basic Experiment}
\begin{figure}[t]
  \centering
  \includegraphics[width=0.85\columnwidth]{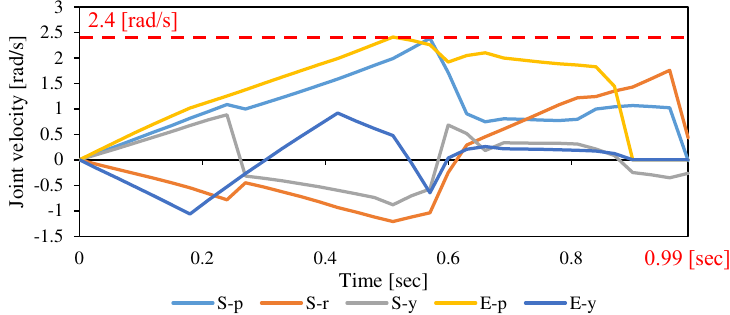}
  \caption{Transition of joint angle velocity in simulation, without any proposed controls.}
  \label{figure:basic-simulation}
  \vspace{-1.0ex}
\end{figure}

\begin{figure}[t]
  \centering
  \includegraphics[width=0.85\columnwidth]{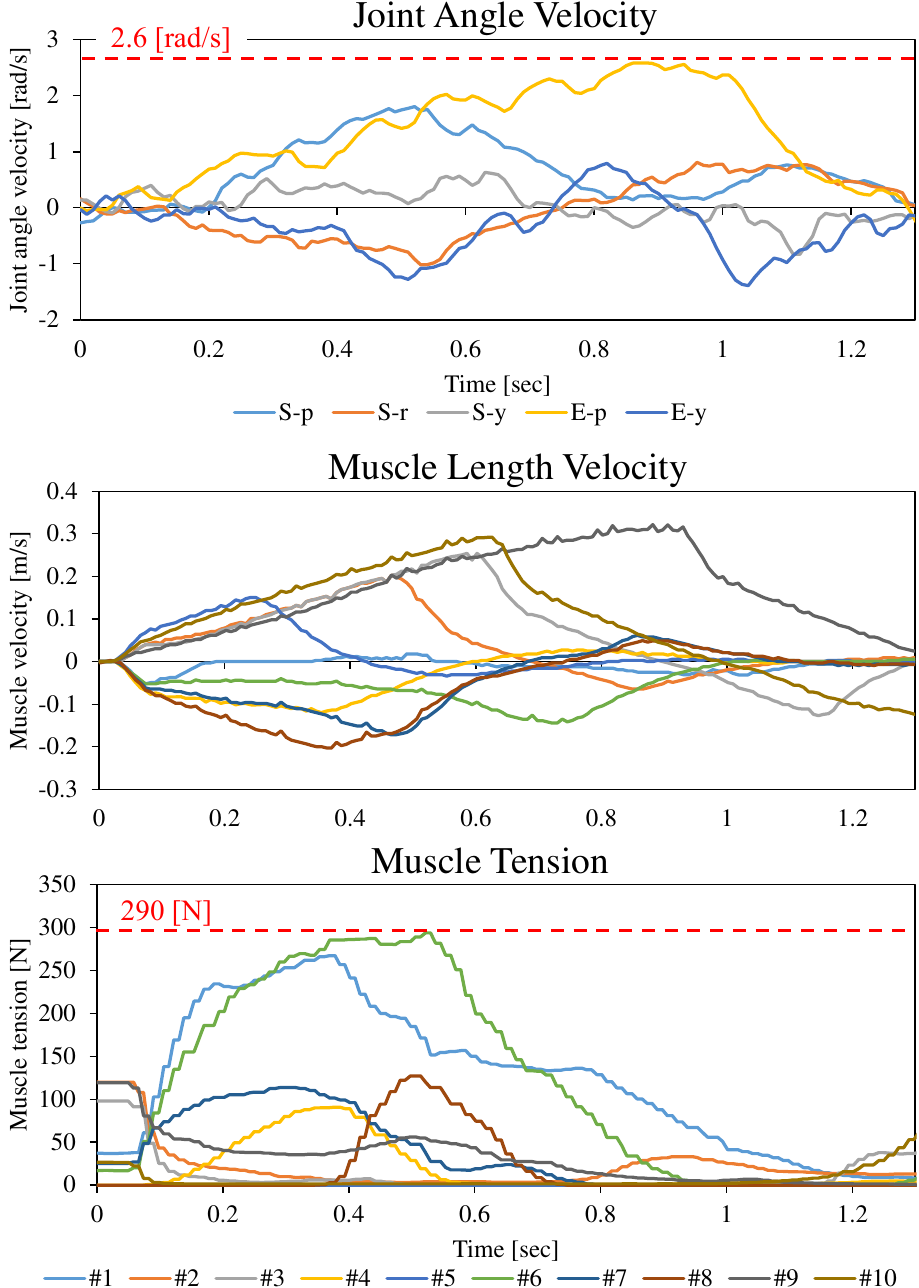}
  \caption{Transition of joint angle velocity, muscle length velocity, and muscle tension in the actual robot, without any proposed controls.}
  \label{figure:basic-actual}
  \vspace{-3.0ex}
\end{figure}

\switchlanguage%
{%
  Before verifying the proposed methods, we conducted the target movement in simulation and in the actual robot without any proposed controls.
  Regarding all experiments, starting from $\bm{\theta}^{start}$, we sent the muscle length achieving $\bm{\theta}^{end}$ for 0 seconds.
  First, we conducted the simulation method explained in the latter half of \secref{subsec:proposed-elongation}.
  We set the mask $\bm{m}$ as a vector whose elements are all 1.
  We show the transition of the joint angle velocity in \figref{figure:basic-simulation}.
  The maximum joint angle velocity was 2.4 rad/s of E-P, and $t^{cost}$ was 0.99 seconds.

  Second, we show the transition of joint angle velocity, muscle length velocity, and muscle tension when conducting the actual robot experiment, in \figref{figure:basic-actual}.
  The maximum joint angle velocity was 2.6 rad/s of E-P, and the result was similar to the simulation.
  The muscle length velocities of the biceps brachii \#9 and brachialis \#10 achieved $\dot{l}^{limit}$.
  Also, the maximum muscle tension was about 290 N, and a heavy load was mainly applied to the agonist muscle of shoulder \#1 and elbow \#6.
  Regarding antagonist muscles, about 50 N was constantly applied to the polyarticular muscle of the biceps brachii \#9.
}%
{%
  提案手法を検証する前に, シミュレーション・実機において本動作ではどの程度の関節角速度・筋長速度が出るのかを確認する.
  まず, \secref{subsec:proposed-elongation}の後半で説明した方法を用いてシミュレーションを行う.
  このとき, マスク$\bm{m}$は全ての要素が1のベクトルとする.
  そのときの関節角速度遷移を\figref{figure:basic-simulation}に示す.
  最大関節角速度はE-pの2.4 rad/sであり, 収束までに0.99秒かかった.

  次に実機において同じ動作を行った際の関節角速度・筋長速度・筋張力の遷移を\figref{figure:basic-actual}に示す.
  最大関節角速度はE-pの2.6 [rad/s]であり, シミュレーションと似た結果が得られた.
  筋長速度は主に, 拮抗筋である\#9の上腕二頭筋, \#10の上腕筋の速度が上限である$\dot{l}^{limit}$まで達していることがわかる.
  また, このときの筋張力は最大で約290 Nであり, 主に\#1の肩の主動筋, \#6の肘の主動筋に大きな負荷がかかっている.
  拮抗筋としては, \#9の多関節筋である上腕二頭筋に約50 Nの力がかかり続けている.
  なお, 3回動作を行った際にセンサ値の遷移に違いはほとんど見られなかった.
}%

\subsection{Experiment with Method Inhibiting Antagonist Muscles}
\begin{figure}[t]
  \centering
  \includegraphics[width=0.85\columnwidth]{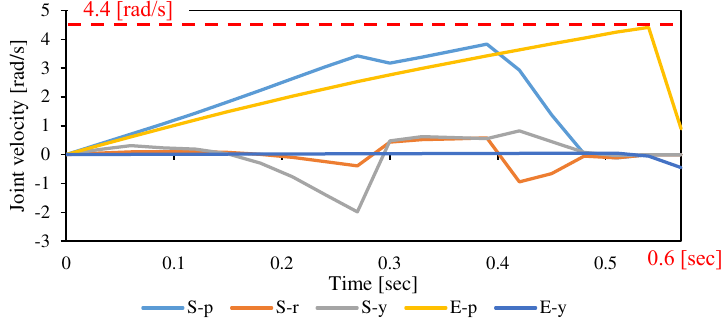}
  \caption{Transition of joint angle velocity when using a method inhibiting antagonist muscles in simulation.}
  \label{figure:backdrivability-simulation}
  \vspace{-1.0ex}
\end{figure}

\begin{figure}[t]
  \centering
  \includegraphics[width=0.85\columnwidth]{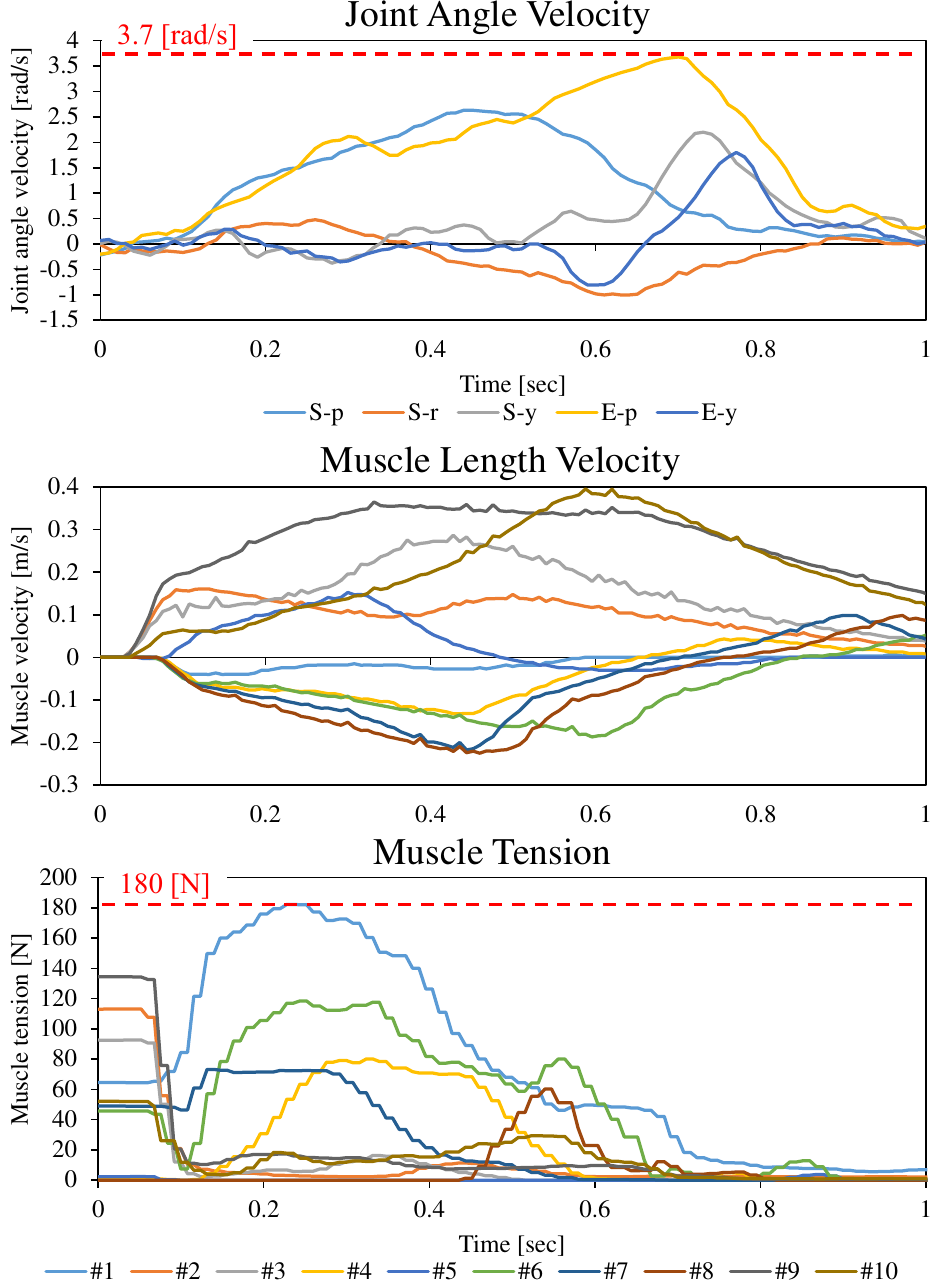}
  \caption{Transition of joint angle velocity, muscle length velocity, and muscle tension when using a method inhibiting antagonist muscles in the actual robot.}
  \label{figure:backdrivability-actual}
  \vspace{-3.0ex}
\end{figure}

\switchlanguage%
{%
  We conducted the target movement in simulation and in the actual robot with the method of \secref{subsec:proposed-backdrivability}.
  First, we conducted a simulation by assuming that the robot has backdrivability and setting the mask $\bm{m}$ as a vector whose elements of muscles with $q[i]>C$ (\#2, \#3, \#9, \#10) are 0 as explained in \equref{eq:backdrivability}.
  We show the transition of the joint angle velocity in \figref{figure:backdrivability-simulation}.
  The maximum joint angle velocity was 4.4 rad/s of E-P, and $t^{cost}$ was 0.6 seconds.

  Second, we show the transition of joint angle velocity, muscle length velocity, and muscle tension when conducting the actual robot experiment, in \figref{figure:backdrivability-actual}.
  The maximum joint angle velocity was 3.7 rad/s of E-P, and the observed velocity was lower than the simulation result.
  The muscle length velocities of the biceps brachii \#9 and brachialis \#10 were faster than $\dot{l}^{limit}$.
  Also, the maximum muscle tension was about 180 N, and a heavy load was mainly applied to the agonist muscle of shoulder \#1 and elbow \#6.
  Regarding antagonist muscles, almost no muscle tension was observed.
}%
{%
  \secref{subsec:proposed-backdrivability}で示した制御をシミュレーション・実機において検証する.
  まずシミュレーションでは, ロボットにbackdrivabilityがあると仮定し, \equref{eq:backdrivability}のように$q[i]>C$となる筋(\#2, \#3, \#9, \#10)のマスク$\bm{m}$を0として動作を行った.
  そのときの関節角速度遷移を\figref{figure:backdrivability-simulation}に示す.
  最大関節角速度はE-pの4.4 rad/sであり, 収束までに0.6秒かかった.

  次に実機において同じ動作を行った際の関節角速度・筋長速度・筋張力の遷移を\figref{figure:backdrivability-actual}に示す.
  最大関節角速度はE-pの3.7 [rad/s]であり, シミュレーションよりは小さめの関節角速度が得られた.
  筋長速度は主に, 拮抗筋である\#9の上腕二頭筋, \#10の上腕筋の速度が上限である$\dot{l}^{limit}$以上の速度を出していることがわかる.
  また, このときの筋張力は最大で約180 Nであり, 主に\#1の肩の主動筋, \#6の肘の主動筋に大きな負荷がかかっている.
  拮抗筋にはほどんと力はかかっていない.
  なお, 3回動作を行った際にセンサ値の遷移に違いはほとんど見られなかった.
}%

\subsection{Experiment with Method Elongating Antagonist Muscles}
\begin{figure}[t]
  \centering
  \includegraphics[width=0.85\columnwidth]{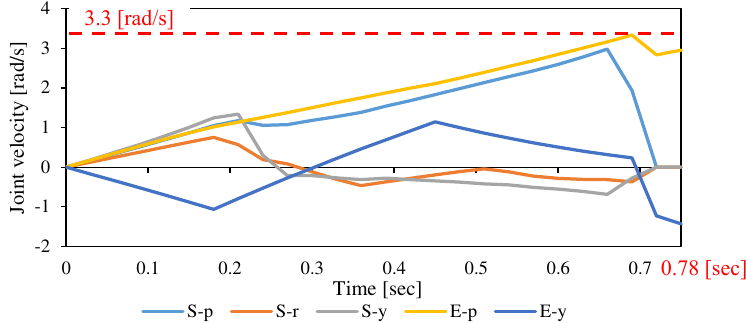}
  \caption{Transition of joint angle velocity when using a method elongating antagonist muscles in simulation.}
  \label{figure:elongation-simulation}
  \vspace{-1.0ex}
\end{figure}

\begin{figure}[t]
  \centering
  \includegraphics[width=0.85\columnwidth]{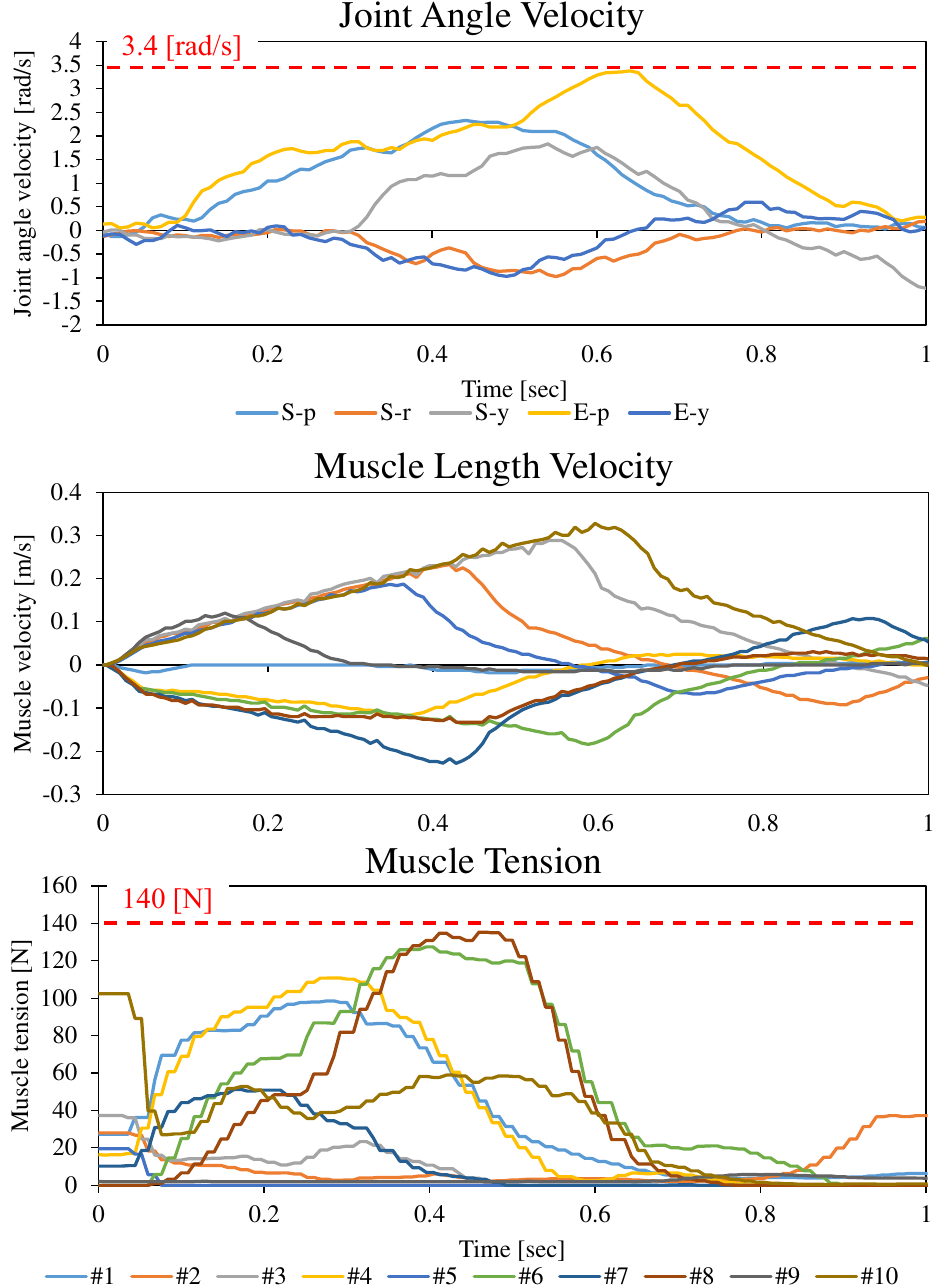}
  \caption{Transition of joint angle velocity, muscle length velocity, and muscle tension when using a method elongating antagonist muscles in the actual robot.}
  \label{figure:elongation-actual}
  \vspace{-3.0ex}
\end{figure}

\switchlanguage%
{%
  We conducted the target movement in simulation and in the actual robot with the method of \secref{subsec:proposed-elongation}.
  First, we calculated the mask $\bm{m}$ satisfying the conditions explained in \secref{subsec:proposed-elongation}.
  Although the $\bm{q}$ of \#9, \#10, and \#3 are large in decreasing order, if both \#9 and \#10 are elongated, the torque of the elbow cannot be kept.
  Therefore, we set $\bm{m}$ as a vector whose element of only \#9 is 0.
  We show the transition of joint angle velocity in \figref{figure:elongation-simulation}.
  The maximum joint angle velocity was 3.3 rad/s of E-P, and $t^{cost}$ was 0.78 seconds.

  Second, we show the transition of the joint angle velocity, muscle length velocity, and muscle tension when conducting the actual robot experiment, in \figref{figure:elongation-actual}.
  The maximum joint angle velocity was 3.4 rad/s of E-P, and the result was similar to the simulation.
  The muscle length velocities of the deltoid (front) \#3 and brachialis \#10 achieved $\dot{l}^{limit}$.
  Also, the maximum muscle tension was about 140 N, and a heavy load was mainly applied to the agonist muscles of elbow \#6 and \#8.
  Regarding antagonist muscles, about 60 N was constantly applied to the brachialis \#10.
}%
{%
  \secref{subsec:proposed-elongation}で示した制御をシミュレーション・実機において検証する.
  まずシミュレーションでは, \secref{subsec:proposed-elongation}に示した方法で, 条件を満たすマスク$\bm{m}$を計算する.
  その結果, モーメントアームが大きな筋は順に\#9, \#10, \#3となるが, \#9と\#10の両者を伸ばしてしまうと, 肘のトルクを確保できないため, \#9のみがマスクされた.
  そのときの関節角速度遷移を\figref{figure:elongation-simulation}に示す.
  最大関節角速度はE-pの3.3 rad/sであり, 収束までに0.78秒かかった.

  次に実機において同じ動作を行った際の関節角速度・筋長速度・筋張力の遷移を\figref{figure:elongation-actual}に示す.
  最大関節角速度はE-pの3.4 [rad/s]であり, シミュレーションとほぼ同じ関節角速度が得られた.
  筋長速度は主に, 拮抗筋である\#3の三角筋前部, \#10の上腕筋の速度が上限である$\dot{l}^{limit}$付近に達していることがわかる.
  また, このときの筋張力は最大で約140 Nであり, 主に\#6, \#8の肘の主動筋に大きな負荷がかかっている.
  拮抗筋としては, \#10の上腕筋に約60 Nの力がかかり続けている.
  なお, 3回動作を行った際にセンサ値の遷移に違いはほとんど見られなかった.
}%

\begin{table}[htb]
  \centering
  \caption{Comparison among the basic motion (\textbf{Basic}), the motion when using the method of \secref{subsec:proposed-backdrivability} (\textbf{Method-1}), and the motion when using the method of \secref{subsec:proposed-elongation} (\textbf{Method-2}).}
  \begin{tabular}{l|ccc}
    & Basic & Method-1 & Method-2 \\ \hline
    Maximum $\dot{\theta}$ (simulation) [rad/s] & 2.4 & 4.4 & 3.3 \\
    Maximum $\dot{\theta}$ (actual robot) [rad/s] & 2.6 & 3.7 & 3.4 \\
    Maximum $\dot{l}$ [m/s] & $\leq\dot{l}^{limit}$ & $>\dot{l}^{limit}$ & $\leq\dot{l}^{limit}$\\
    Maximum $T$ [N] & 290 & 180 & 140
  \end{tabular}
  \label{table:discussion}
  \vspace{-3.0ex}
\end{table}

\section{Discussion} \label{sec:discussion}
\switchlanguage%
{%
  We show the comparison among the ordinary movement (\textbf{Basic}) and movements using the method of \secref{subsec:proposed-backdrivability} (\textbf{Method-1}) or \secref{subsec:proposed-elongation} (\textbf{Method-2}), in \tabref{table:discussion}.
  First, from the simulation results, the theoretical maximum joint angle velocity has the relationship of \textbf{Basic}$<$\textbf{Method-2}$<$\textbf{Method-1}.
  Also, the relationship of the actual robot experiments is the same with that of the simulation.
  Thus, the methods of this study are effective in maximizing joint angle velocity.
  However, regarding \textbf{Method-1}, there is a large error between the simulation and actual robot experiments.
  This is becase \textbf{Method-1} is a method that assumes the muscle actuators have backdrivability.
  Although the gear ratio of the muscle actuator is relatively low, 29:1, the joint angle velocity can decrease if we increase the gear ratio, due to the lack of backdrivability.
  Second, we consider the difference of muscle length velocities.
  Because \textbf{Method-1} makes the antagonist muscles elongate spontaneously, the muscle length velocity is higher than $\dot{l}^{limit}$.
  On the other hand, because \textbf{Method-2} elongates the antagonist muscles in advance, we cannot see the high muscle length velocity.
  Third, we consider the difference of muscle tensions.
  While large muscle tension emerges by large internal force regarding \textbf{Basic}, only about half of the muscle tension emerges regarding \textbf{Method-1} and \textbf{Method-2}.
  Thus, by inhibiting antagonist muscles or elongating them in advance, not only is joint angle velocity maximized but also muscle tension is reduced.

  Summarizing the above, although \textbf{Method-1} is effective if the backdrivability is high, the performance can be worse than \textbf{Method-2} if the backdrivability is low.
  On the other hand, while the performance of \textbf{Method-2} is usually worse than that of \textbf{Method-1}, \textbf{Method-2} does not depend on the backdrivability.
  However, because \textbf{Method-2} elongates antagonist muscles in advance, high muscle tension is necessary when the joint angle is $\bm{\theta}^{start}$.

  In this study, we developed simple methods exceeding the limited maximum joint angle velocity.
  By modelizing the friction, hysteresis, and dynamics better, we can analyze the performance in more detail.
  In the future, we need to develop a method of realizing the accurate joint angle trajectory with fast velocity.
}%
{%
  通常の動作(\textbf{Basic})と\secref{subsec:proposed-backdrivability}(\textbf{Method-1}), \secref{subsec:proposed-elongation}(\textbf{Method-2})を用いたときの動作の比較を\tabref{table:discussion}に示す.
  まず, シミュレーションから理論的な最大関節角速度の大きさは\textbf{Basic}$<$\textbf{Method-2}$<$\textbf{Method-1}であることがわかった.
  また実機実験より, その結果は実機に置いても同様であることが示された.
  よって, 本研究の手法は関節角速度最大化において有効である.
  しかし, \textbf{Method-1}においては, シミュレーションと実機で大きな誤差がある.
  これは, \textbf{Method-1}がバックドライバビリティを前提とした手法であることが理由として挙げられる.
  そのため, 現在のアクチュエータのギア比は29:1であるが, これをより高くしてしまうと, 関節角速度はさらに落ちる可能性がある.
  次に, 筋長速度の違いについて考察する.
  \textbf{Method-1}では拮抗筋はバックドライバビリティによって勝手に伸びるため, $\dot{l}^{limit}$よりも大きな速度が出ていることがわかる.
  それに対して, \textbf{Method-2}は最初から拮抗筋を伸ばしてしまうため, 大きな筋長速度を出さずに済んでいることがわかる.
  最後に, 筋張力について考察する.
  \textbf{Basic}では強い拮抗によって大きな力が発揮されているのに対して, \textbf{Method-1/2}では半分程度の筋張力しか発揮されていない.
  よって, 拮抗筋を抑制, または予め伸ばすことで, 関節角速度最大化だけでなく, 筋張力の削減にも効果があると考えられる.

  まとめると, \textbf{Method-1}はバックドライバビリティが高いならば有効であるが, バックドライバビリティが低い場合は\textbf{Method-2}よりも性能が悪くなる可能性がある.
  これに対して, \textbf{Method-2}は\textbf{Method-1}に比べて性能が劣ることがある反面, バックドライバビリティに依存せずに有効である.
  しかし, 予め筋を伸ばすため, 関節角度が$\bm{\theta}^{start}$のときに高い筋張力が必要になってしまうことがある.

  本研究は非常にシンプルな手法によってより大きな関節角速度を出す手法を開発した.
  しかし, ロボットの摩擦やヒステリシス等の動的な要素や, より良いモデル化によって, より詳しい解析ができる可能性がある.
  また今後, 速い動きをしたときの, 正確な関節角度遷移の実現も必要になっていくと考える.
}%

\section{CONCLUSION} \label{sec:conclusion}
\switchlanguage%
{%
  In this study, we proposed two methods to exceed the maximum joint angle velocity limited by the actuator specifications for musculoskeletal humanoids with redundant tendon-driven structures.
  One of them is a method inhibiting antagonist muscles, thus making their current 0 and using backdrivability of muscles.
  Another one is a method elongating a few of the antagonist muscles in advance.
  From the simulation and actual robot experiments, we verified that the two methods can work well.
  Also, the performance of the former depends on the backdrivability, and that of the latter does not.

  In future works, we would like to apply this method to more realistic situations.
}%
{%
  本研究では, 冗長で複雑な拮抗腱駆動構造を有する筋骨格ヒューマノイドにおいて, アクチュエータによって規定される最大関節角速度よりも大きな速度を出す手法を2つ提案した.
  一つは動作時に拮抗筋の電流値を0として抑制し, バックドライバビリティを用いる手法である.
  もう一つは動作前に予め弛緩可能な拮抗筋を弛緩させてしまう手法である.
  両者ともシミュレーション・実機実験からより速い関節角速度を出すことが可能であることがわかった.
  また, 前者はバックドライバビリティに依存し, 後者はバックドライバビリティに依存しない.

  今後は, 本手法の現実タスクへの応用について考察をしていきたい.
}%

{
  \bibliographystyle{IEEEtran}
  \bibliography{main}
}

\end{document}